\pdfoutput=1

\documentclass[11pt,a4paper]{article}
\usepackage[hyperref]{acl2021}
\usepackage{url}            
\PassOptionsToPackage{hyphens}{url}
\usepackage{times}
\usepackage{latexsym}
\usepackage{graphicx}
\usepackage{tabularx}
\usepackage{booktabs}
\usepackage{amsmath}
\usepackage{bm}
\usepackage{amssymb}
\usepackage{enumitem}
\usepackage{multirow}
\usepackage{color, colortbl}
\definecolor{Gray}{gray}{0.9}

\usepackage{microtype}

\usepackage{flushend}   
\aclfinalcopy 
\def\aclpaperid{3374} 

\newcommand\minilm{\textsc{MiniLM}}
\newcommand\prophetnet{ProphetNet}
\newcommand\meqsum{\textsc{MeQSum}}
\newcommand\matinf{\textsc{MATINF}}
\newcommand\bertbase{BERT$_\text{BASE}$}

\title{Reinforcement Learning for Abstractive Question Summarization with Question-aware Semantic Rewards }

\author{
Shweta Yadav\thanks{$^*$These authors contributed equally to this work.}, \hspace{0.1cm} Deepak Gupta\footnotemark[1], \hspace{0.1cm} Asma Ben Abacha, \hspace{0.1cm} Dina Demner-Fushman  \\
   LHNCBC, U.S. National Library of Medicine, MD, USA  \\
  {\tt 
  \{shweta.shweta}, 
    {\tt 
  deepak.gupta},
      {\tt 
  asma.benabacha\}@nih.gov}\\
  {\tt ddemner@mail.nih.gov
  }
  } 
\date{}

\begin{document}
\maketitle

\begin{abstract}
The growth of online consumer health questions has led to the necessity for reliable and accurate question answering systems. A recent study showed that manual summarization of consumer health questions brings significant improvement in retrieving relevant answers. However, the automatic summarization of long questions is a challenging task due to the lack of training data and the complexity of the related subtasks, such as the question focus and type recognition. In this paper, we introduce a reinforcement learning-based framework for abstractive question summarization. We propose two novel rewards obtained from the downstream tasks of (i) question-type identification and (ii) question-focus recognition to regularize the question generation model. These rewards ensure the generation of semantically valid questions and encourage the inclusion of key medical entities/foci in the question summary. We evaluated our proposed method on two benchmark datasets and achieved higher performance over state-of-the-art models. The manual evaluation of the summaries reveals that the generated questions are more diverse and have fewer factual inconsistencies than the baseline summaries. 
The source code is available here: \url{https://github.com/shwetanlp/CHQ-Summ}.   

\end{abstract}

\section{Introduction} \label{sec:intro}
The growing trend in online web forums is to attract more and more consumers to use the Internet for their health information needs. An instinctive way for consumers to query for their health-related content is in the form of natural language questions.  
These questions are often excessively descriptive and contain more than required peripheral information.
However, most of the textual content is not particularly relevant in answering the question \cite{kilicoglu2013interpreting}. A recent study showed that manual summarization of consumer health questions (CHQ) has significant improvement (58\%) in retrieving relevant answers \cite{abacha2019role}. 
However, three major limitations impede higher success in obtaining semantically and factually correct summaries: \textbf{(1)} the complexity of identifying the correct question type/intent, \textbf{(2)} the difficulty of identifying salient medical entities and focus/topic of the question, and \textbf{(3)} the lack of large-scale CHQ summarization datasets. 
To address these limitations, this work presents a new reinforcement learning based framework for abstractive question summarization. We also propose two novel question-aware semantic reward functions: Question-type Identification Reward (QTR) and Question-focus Recognition Reward (QFR). The QTR measures correctly identified question-type(s) of the summarized question. Similarly, QFR measures correctly recognized key medical concept(s) or focus/foci of the summary.   

We use the reinforce-based policy gradient approach, which maximizes the non-differentiable QTR and QFR rewards by learning the optimal policy defined by the Transformer model parameters. Our experiments show that these two rewards can significantly improve the question summarization quality, separately or jointly, achieving the new state-of-the-art performance on the \meqsum{} and \matinf{} benchmark datasets. The main  contributions of this paper are as follows:
\vspace{-2mm}
\begin{itemize}[noitemsep]
\item We propose a novel approach towards question summarization by introducing two question-aware semantic rewards \textbf{(i)} \textit{Question-type Identification Reward} and \textbf{(ii)} \textit{Question-focus Recognition Reward}, to enforce the generation of semantically valid and factually correct question summaries. 
\item The proposed models achieve the state-of-the-art performance on two question summarization datasets over competitive pre-trained Transformer models. 
\item  A manual evaluation of the summarized questions reveals that they achieve higher abstraction levels and are more semantically and factually similar to human-generated summaries.
\end{itemize}

\section{Related Work} \label{sec:related} 
 In recent years, reinforcement learning (RL) based models have been explored for the abstractive summarization task. \citet{paulus2017deep} introduced RL in neural summarization models by optimizing the ROUGE score as a reward that led to more readable and concise summaries. Subsequently, several studies \cite{chen2018fast,pasunuru2018multi,zhang-bansal-2019-addressing, gupta2020reinforced,zhang2019optimizing} have proposed methods to optimize the model losses via RL that enables the model to generate the sentences with the higher  ROUGE score.  
 While these methods are primarily supervised, \citet{laban2020summary} proposed an unsupervised method that accounts for fluency, brevity, and coverage in generated summaries using multiple RL-based rewards. 
 The majority of these works are focused on document summarization with conventional non-semantics rewards (ROUGE, BLEU). In contrast, we focus on formulating the semantic rewards that bring a high-level semantic regularization. In particular, we investigate the question's main characteristics, i.e., question focus and type, to define the rewards.\\
\indent Recently, \citet{BenAbacha-ACL-2019-Sum} defined the CHQ summarization task and introduced a new benchmark (\meqsum{}) and a pointer-generator model. \citet{ben-abacha-etal-2021-overview} organized the MEDIQA-21 shared task challenge on CHQ, multi-document answers, and radiology report summarization. Most of the participating team \cite{yadav-etal-2021-nlm,he-etal-2021-damo,sanger-etal-2021-wbi} utilized transfer learning, knowledge-based, and ensemble methods to solve the question summarization task.
\citet{yadav2021questionaware} proposed  question-aware transformer models for question summarization.
\citet{xu2020matinf} automatically created a Chinese dataset (MATINF) for medical question answering, summarization, and classification tasks focusing on maternity and infant categories. 
Some of the other prominent works in the abstractive summarization of long and short documents include \citet{cohan2018discourse,zhang2019pegasus,macavaney2019ontology,sotudeh2020attend}.

\section{Proposed Method} \label{sec:method}
Given a question, the goal of the task is to generate a summarized question that contains the salient information of the original question. We propose a RL-based question summarizer model over the Transformer \cite{vaswani2017attention} encoder-decoder architecture. We describe below the proposed reward functions. 



\subsection{Question-aware Semantic Rewards} \label{sec:rewards}
\textbf{(a) Question-type Identification Reward: } 
Independent of the pre-training task, most language models use maximum likelihood estimation (MLE)-based training for fine-tuning the downstream tasks. MLE has two drawbacks: (1) ``\textit{exposure bias}" \cite{exposure-bias} when the model expects gold-standard data at each step during training but does not have such supervision when testing, 
 and (2) ``\textit{representational collapse}'' \cite{aghajanyan2021better}, is the degradation of generalizable
representations of pre-trained models during the fine-tuning stage. 
To deal with the \textit{exposure bias}, previous works used the ROUGE and BLEU rewards to train the generation models \cite{paulus2017deep,exposure-bias}. These evaluation metrics are based on n-grams matching and might fail to capture the semantics of the generated questions. We, therefore, propose a new question-type identification reward to capture the underlying question semantics. 

We fine-tuned a \bertbase{} network as a question-type identification model to provide question-type labels. Specifically, we use the \texttt{[CLS]} token representation ($\bm{h}_{[CLS]}$) from the final transformer layer of \bertbase{} and add the feed-forward layers on top of the $\bm{h}_{[CLS]}$ to compute the final logits 
\begin{center}
    $l = \bm{W}(tanh(\bm{U}h_{[CLS]} + \bm{a})) + \bm{b}$
\end{center} 
Finally, the question types are predicted using the \textit{sigmoid} activation function on each output neuron of logits $l$. The fine-tuned network is used to compute the reward $r_{QTR}(Q^p, Q^*)$ as F-Score of question-types between the generated question summary $Q^p$ and the gold question summary $Q^*$.


\paragraph{\textbf{(b) Question-focus Recognition Reward: }} A good question summary should contain the key information of the original question to avoid factual inconsistency. In the literature, ROUGE-based rewards have been explored to maximize the coverage of the generated summary, but it does not guarantee to preserve the key information in the question summary. We introduce a novel reward function called question-focus recognition reward, which captures the degree to which the key information from the original question is present in the generated summary question. Similar to QTR, we fine-tuned the \bertbase{} network for question-focus recognition to predict the focus/foci of the question. Specifically, given the representation matrix ($\bm{H} \in \mathcal{R}^{n \times d} $) of $n$ tokens and $d$ dimensional hidden state representation obtained from the final transformer layer of \bertbase{}, we performed the token level prediction using a linear layer of the feed-forward network. For each token representation ($h_i$), we compute the logits $l_i \in \mathcal{R}^{|C|}$, where ($|C|$) is the number of classes and predict the question focus as follows: $f_i= softmax(\bm{W}h_i + \bm{b}) $. 
The fine-tuned network is used to compute the reward $r_{QFR}(Q^p, Q^*)$ as F-Score of question-focus between the generated question summary $Q^p$ and the gold question summary $Q^*$. 
\subsection{Policy Gradient REINFORCE}
We cast question summarization as an RL problem, where the ``\textit{agent}'' (ProphetNet decoder) interacts with the ``\textit{environment}'' (Question-type or focus prediction networks) to take ``\textit{actions}'' (next word prediction) based on the learned ``policy'' $p_\theta$ defined by ProphetNet parameters ($\theta$) and observe ``\textit{reward}'' (QTR and QFR). 
We utilized \prophetnet{} \cite{qi-etal-2020-prophetnet} as the base model because it is specifically designed for sequence-to-sequence training and it has shown near state-of-the-art results on natural language generation task.
We use the REINFORCE algorithm \cite{williams1992simple} to learn the optimal policy which maximizes the expected reward. Toward this, we minimize the loss function $\mathcal{L}_{RL} = -E_{Q^s \sim p_\theta}[r(Q^s, Q^*)]$, where $Q^s$ is the question formed by sampling the words $q_{t}^{s}$ from the model's output distribution, i.e. $p(q_t^s|q_1^s,  q_2^s, \ldots, q_{t-1}^s, \mathcal{S})$. The derivative of $\mathcal{L}_{RL}$ is approximated using a single sample along with baseline estimator $b$:
\begin{equation}
    \bigtriangledown_\theta \mathcal{L}_{RL}  = -(r(Q^s, Q^*) - b)\bigtriangledown_\theta log p_\theta(Q^s)
\end{equation}
The Self-critical Sequence Training (SCST) strategy \cite{rennie2017self} is used to estimate the baseline reward by computing the reward with the question generated by the current model using the greedy decoding technique, i.e., $b = r(Q^g, Q^*)$. 
We compute the final reward as a weighted sum of QTR and QFR as follows:
\begin{equation}
\footnotesize
    r(Q^p, Q^*) = \gamma_{QTR} \times r_{QTR}(Q^p, Q^*) + \gamma_{QFR} \times r_{QFR}(Q^p, Q^*)
\end{equation}
We train the network with the mixed loss as discussed in  \citet{paulus2017deep}. The overall network loss is as follows:
\begin{equation}
\label{main-eq}
    \mathcal{L}  = \alpha \mathcal{L}_{RL}  + (1-\alpha)\mathcal{L}_{ML}
\end{equation}
where, $\alpha$ is the scaling factor and $\mathcal{L}_{ML}$ is the negative log-likelihood loss and equivalent to $-\sum_{t=1}^{t=m} logp(q_t^*|q_1^*,  q_2^*, \ldots, q_{t-1}^*, \mathcal{S})$, where $\mathcal{S}$ is the source question.
\section{Experimental Results \& Analysis}
\subsection{Datasets}  
We utilized two CHQ abstractive summarization datasets: \meqsum{} and \matinf{}\footnote{Since the dataset was in Chinese, we translated it to English using Google Translate.} to evaluate the proposed framework. The \meqsum{}\footnote{\url{https://github.com/abachaa/MeQSum}} training set consists of $5,155$ CHQ-summary pairs and the test set includes $500$ pairs. We chose $100$ samples from the training set as the validation dataset. \\
\indent For fine-tuning the question-type identification and question-focus recognition models, we manually labeled the \meqsum{} dataset with the question type: (\textit{`Dosage', `Drugs', `Diagnosis', `Treatments', `Duration', `Testing', `Symptom', `Usage', `Information', `Causes'}) and foci. We use the labeled data to train the question-type identification and question-focus recognition networks. For question-focus recognition, we follow the BIO notation and classify each token for the beginning  of focus token (\textbf{B}), intermediate of focus token (\textbf{I}), and other token (\textbf{O}) classes. Since, the gold annotations for question-types and question-focus were not available for the \matinf{} dataset, we used the pre-trained network trained on the \meqsum{} dataset to obtain the silver-standard question-types and question-focus information for \matinf{}\footnote{\url{https://github.com/WHUIR/MATINF}}.
The \matinf{}{} dataset has $5,000$ CHQ-summary pairs in the training set and $500$ in the test set. 

\subsection{Experimental Setups}
\begin{table*}[]
\centering
\resizebox{0.85\linewidth}{!}{%
\begin{tabular}{l|l|ccc|ccc}
 \hline
 \multirow{2}{*}{}  &
\multirow{2}{*}{\textbf{Models}} & \multicolumn{3}{c}{\meqsum{}} & \multicolumn{3}{c}{\matinf{}$^*$} \\ \cline{3-8} 
    &   & \textbf{R-1 }     & \textbf{R-2} & \textbf{R-L}   &  \textbf{R-1 } & \textbf{R-2}     & \textbf{R-L}
 \\ \hline 
 \multirow{11}{*}{\rotatebox[origin=c]{90}{\textbf{Baselines}}} &\begin{tabular}[c]{@{}c@{}} Seq2Seq \cite{seq2seq}\\  \end{tabular}    &25.28  &14.39  &24.64 & $17.77$ & $5.10$ & $21.48$\\ 
 &Seq2Seq + Attention \cite{bahadanau_attention}    
 & 28.11    & $17.24$   &$27.82$ & $19.45$ & $6.45$ & $23.77$ \\
  &Pointer Generator (PG) \cite{go-to-the-point-see-2017}   & 32.41  & 19.37  & 36.53 & 23.31 & 7.01 & 26.61\\ 
  &\begin{tabular}[c]{@{}c@{}}SOTA \cite{BenAbacha-ACL-2019-Sum}  \end{tabular}  &  44.16 & {27.64}  & 42.78 & $-$ & $-$ & $-$ \\ 
 &\begin{tabular}[c]{@{}c@{}}SOTA$^*$ \cite{BenAbacha-ACL-2019-Sum}  \end{tabular}  & 40.00 & 24.13  & 38.56 & $24.58$ & $7.30$ & $28.08$\\ \cline{2-8}
 
 &Transformer \cite{vaswani2017attention}  & 25.84   & 13.66   & 29.12 &22.25 &5.89 & 26.06\\
 &BertSumm \cite{liu-lapata-2019-text}  & 26.24   & 16.20   & 30.59  &31.16 & 11.94 &34.70\\ 
 &T5$_{\text{BASE}}$ \cite{raffel2019exploring}  & 38.92   & 21.29   & 40.56 & 39.66 &21.24 &41.52\\ 
 &PEGASUS \cite{zhang2019pegasus} & 39.06   & 20.18   & 42.05 & 40.05 &23.67 & 43.30 \\
 &BART$_{\text{LARGE}}$ \cite{lewis2019bart}   & 42.30   & 24.83   & 43.74 & 42.52 & 23.13 & 43.98  \\ 
 &\minilm{} \cite{wang2020minilm}  & 43.13   & 26.03   & 46.39 &35.60 &18.08 &38.70\\
 & \prophetnet{} \cite{qi-etal-2020-prophetnet}  & 43.87   & 25.99   & 46.52 & 46.94 & 27.77 & 48.43 \\
  & \prophetnet{} + ROUGE-L  & 44.33 & 26.32  & 46.90 & 48.17 & 28.13 & 48.66
  \\ \hline
\hline
 \multirow{3}{*}{\rotatebox[origin=c]{90}{\footnotesize \textbf{\begin{tabular}[c]{@{}c@{}}Joint\\ Learning\end{tabular}}}} 
 &\prophetnet{} + Q-type  & 44.40  & 26.63  &47.05 & 47.19 & 28.02 & 48.70\\ 
  & \prophetnet{} + Q-focus  & 44.62  & 26.61  &47.28 & 47.14 & 28.06 & 48.64  \\ 
  &  \prophetnet{} + Q-type + Q-focus  & 44.67  & 26.72  &47.34 & 47.18 & 28.04 & 48.65  \\ \hline
  \hline
 \multirow{3}{*}{\rotatebox[origin=c]{90}{\footnotesize \textbf{\begin{tabular}[c]{@{}c@{}}Proposed \\ Approach\end{tabular}}}} 
 &\prophetnet{} + QTR  & 44.60  & 26.69  &47.38 & 47.51 & 28.40 & 48.94\\ 
  & \prophetnet{} + QFR  &45.36   &27.33   &47.96  & 47.53 &28.29  &49.11  \\ 
  &  \textbf{\prophetnet{} + QTR + QFR}  & {45.52}  & {27.54}  &\textbf{48.19} & {47.73} & {28.54} & \textbf{49.33}  \\ \hline
\end{tabular} 
} 
\caption{Comparison of the proposed models and various baselines. SOTA$^*$ denotes the method trained on the same data that we used. \matinf{}$^*$ denotes a translated English subset of the original Chinese \matinf{} dataset. 
} 
\label{tab:result}
\end{table*} 
\begin{table*}[!h]
\centering
\resizebox{0.8\linewidth}{!}{%
\begin{tabular}{l|cccc|cccc} 
\hline
        \multirow{2}{*}{\textbf{Summary Label}} &         \multicolumn{4}{c|}{\meqsum{}} & \multicolumn{4}{c}{\matinf{}} \\ \cline{2-9}
                &
  \textbf{M1}   & \textbf{M2}   & \textbf{M3}  & \textbf{M4}  & \textbf{M1}   & \textbf{M2}   & \textbf{M3}  & \textbf{M4}  \\ \hline
Semantics Preserved (PC/FC) &14/19.5      &9.5/29      &18/28     &  19.5/29   & 6/32.5     &  9.5/33    &  13.5/34   & 14/35     \\ 
Factual Consistent (PC/FC) &11/25      & 7.5/35     &9.5/36.5  &  10/38 & 5.5/35    & 7/36     & 7.5/41     & 9/42.5      \\ \hline
Incorrect       &  23    &11      &12.5     & 11     & 10.5     & 11.5     & 11.5    & 10     \\ 
Acceptable      &  18.5    & 10     & 12.5    & 12.5     &  15   & 10.5     &  8.5   &  9.5    \\ 
Perfect         &   8.5   & 29     &  25   & 26.5    &  24.5    &  28    & 30    & 30.5     \\ \hline
\end{tabular}
} 
\caption{Results of the manual evaluation of the summaries generated by \prophetnet{} (M1), M1+QTR (M2), M1+QFR (M3), and M1+QTR+QFR (M4). For \textit{Semantic Preserved} and \textit{Factual Consistent}, we report the partially correct (PC) and fully correct (FC) numbers.}  
\label{tab:ManualVal}  
\end{table*}

\begin{table}[!h]
\centering
\begin{tabular}{@{}p{\columnwidth}@{}}
\specialrule{.1em}{.1em}{.1em}
\small
\textbf{Original Question-I:} \textcolor{black}{
who makes bromocriptine i am wondering what company makes the drug bromocriptine, i need it for a mass i have on my pituitary gland and the cost just keeps raising. i cannot ever buy a full prescription because of the price and i was told if i get a hold of the maker of the drug sometimes they offer coupons or something to help me afford the medicine. if i buy 10 pills in which i have to take 2 times a day it costs me 78.00. and that is how i have to buy them.  thanks.} \\
\midrule
\small
\textbf{Reference:} 
\textcolor{blue}{who manufactures bromocriptine?} \\
\midrule
\small
\textbf{Generated Summary} \\ \addlinespace[-0.1cm]
\midrule
\small
\textbf{\prophetnet{}:} \textcolor{red}{what is bromocriptine?}
\\ \hline
\addlinespace[-0.1cm]
\small
\textbf{Proposed Approach}: \textcolor{blue}{what company makes bromocriptine and how much does it cost?}\\
\specialrule{.1em}{.1em}{.1em}
\footnotesize
\textbf{Original Question-II:} 
Have been on methadone for four years. I am interested in the rapid withdrawal under anesthesia, but do not have a clue where I can find a doctor or hospital who does this. I also would like to know the approximate cost and if or what insurance companies pay for this.\\
\midrule
\small
\textbf{Reference:}\textcolor{blue}{ 
how can I find a physician (s) or hospital (s) who specialize in rapid methadone withdrawal under anesthesia, and the cost and insurance benefits for the procedure?} \\ 
\midrule
\small
\textbf{Generated Summary} \\ \addlinespace[-0.1cm]
\midrule
\small
\textbf{\prophetnet{}:} \textcolor{red}{ what is the treatment for rapid withdrawal of methadone under anesthesia?}
\\
\addlinespace[-0.15cm]
\small
\textbf{Proposed Approach}: \textcolor{blue}{where can i find physician (s) who specialize in rapid withdrawal of methadone?}\\
\specialrule{.1em}{.1em}{.1em}
\end{tabular}
\caption{Correct/Incorrect summaries generated on \meqsum{}. Example-I shows a perfect summary over \prophetnet{}. The second example shows an incorrect summary with a partially extracted focus (`\textit{under anesthesia}') and two missing types (`\textit{cost}', `\textit{procedures}'). }   
\label{tab:results-analysis}
\vspace{-4mm} 
\end{table}

We use the pre-trained uncased version\footnote{\url{https://huggingface.co/microsoft/prophetnet-large-uncased}} of ProphetNet as the base encoder-decoder model.
We use a beam search algorithm with beam size $4$ to decode the summary sentence. We train all summarization models on the respective training dataset for $20$ epochs. We set the maximum question and summary sentence length to $120$ and $20$,  respectively. We first fine-train the proposed network by minimizing only the maximum likelihood (ML) loss. Next, we initialize our proposed model with the fine-trained ML weights and train the network with the mixed-objective learning function (Eq. \ref{main-eq}).  
We performed experiments on the validation dataset by varying the $\alpha, \gamma_{QTR}$ and $\gamma_{QFR}$ in the range of $(0,1)$.
The scaling factor ($\alpha$) value $0.95$, was found to be optimal (in terms of Rouge-L) for both the datasets. The values of $\gamma_{QTR}=0.4$ and $\gamma_{QFR}=0.6$ were found to be optimal on the validation sets of both datasets. 
To update the model parameters, we used Adam \cite{adam} optimization algorithm with the learning rate of $7e-5$ for ML training and $3e-7 $ for RL training. We obtained the optimal hyper-parameters values based on the performance of the model on the validation sets of \meqsum{} and \matinf{} in the respective experiments.
 We used a cosine annealing learning rate \cite{sgdr} decay schedule, where the learning rate decreases linearly from the initial learning set in the optimizer to $0$. To avoid the gradient explosion issue, the gradient norm was clipped within $1$. For all the baseline experiments, we followed the official source code of the approach and trained the model on our datasets. We implemented the approach of  \citet{BenAbacha-ACL-2019-Sum} to evaluate the performance on both datasets.  All experiments were performed on a single NVIDIA Tesla V100 GPU having GPU memory of $32$GB. The average runtimes (each epoch) for the proposed approaches $M_2$, $M_3$ and $M_4$ were $2.7, 2.8$ and $4.5$ hours, respectively. All the proposed models have $391.32$ million parameters.

\subsection{Results}
We present the results of the proposed question-aware semantic rewards on the \meqsum{} and \matinf{} datasets in Table-\ref{tab:result}. We evaluated the generated summaries using the ROUGE \cite{lin2004rouge} metric\footnote{\url{https://pypi.org/project/py-rouge/}}. The proposed model achieves new state-of-the-art performance on both datasets by outperforming competitive baseline Transformer models. We also compare the proposed model with the joint learning baselines, where we regularize the question summarizer with the additional loss obtained from the question-type (Q-type) identification and question-focus (Q-focus) recognition model. To make a fair comparison with the proposed approach, we train these joint learning-based models with the same weighted strategy shown in Eq. \ref{main-eq}. The results reported in Table \ref{tab:result} show the improvement over the \prophetnet{} on both datasets.\\

\indent In comparison to the benchmark model on \meqsum{}, our proposed model obtained an improvement of $9.63\%$. A similar improvement is also observed on the \matinf{} dataset. 
Furthermore, the results show that individual QTR and QFR rewards also improve over \prophetnet{} and ROUGE-based rewards. These results support two major claims: \textbf{(1)} question-type reward assists the model to capture the underlying question semantics, and \textbf{(2)} awareness of salient entities learned from the question-focus reward enables the generation of fewer incorrect summaries that are unrelated to the question topic. The proposed rewards are model-independent and can be plugged into any pre-trained Seq2Seq model. 
On the downstream tasks of question-type identification and question-focus recognition, the pre-trained BERT model achieves the F-Score of $97.10\%$ and $77.24\%$, respectively, on 10\% of the manually labeled \meqsum{} pairs.

\paragraph{Manual Evaluation:} 
Two annotators, experts in medical informatics, performed an analysis 
of 50 summaries randomly selected from each test set. In \matinf{}, nine out of the 50 samples contained translation errors. We thus randomly replaced them. In both datasets, we annotated each summary with two labels `\textit{Semantics Preserved}' and `\textit{Factual Consistent}' to measure \textbf{(1)} whether the semantics (i.e., question intent) of the source question was preserved in the generated summary and \textbf{(2)} whether the key entities/foci were present in the generated summary.  In the manual evaluation of the quality of the generated summaries, we categorize each summary into one of the following categories: `\textit{Incorrect}',  `\textit{Acceptable}', and `\textit{Perfect}'.
We report the human evaluation results (average of two annotators) on both datasets in Table-\ref{tab:ManualVal}. The results show that our proposed rewards enhance the model by capturing the underlying semantics and facts, which led to higher proportions of perfect and acceptable summaries.
The error analysis identified two major causes of errors: \textbf{(1)} Wrong question types (e.g. the original question contained multiple question types or has insufficient type-related training instances) and \textbf{(2)} Wrong/partial focus (e.g. the model fails to capture the key medical entities).

\section{Conclusion}
In this work, we present an RL-based framework by introducing novel question-aware semantic rewards to enhance the semantics and factual consistency of the summarized questions. The automatic and human evaluations demonstrated the efficiency of these rewards when integrated with a strong encoder-decoder based ProphetNet transformer model. The proposed methods achieve state-of-the-art results on two-question summarization benchmarks. In the future, we will explore other types of semantic rewards and efficient multi-rewards optimization algorithms for RL.     
\section*{Acknowledgements}
This research was supported by the Intramural Research Program of the National Library of Medicine, National Institutes of Health.
\section*{Ethics / Impact Statement}
Our project involves publicly available datasets of consumer health questions. It does not involve any direct interaction with any individuals or their personally identifiable data and does not meet the Federal definition for human subjects research, specifically: ``a systematic investigation designed to contribute to generalizable knowledge" and ``research involving interaction with the individual or obtains personally identifiable private information about an individual."

\bibliographystyle{acl_natbib}
\bibliography{anthology,acl2021}


\end{document}